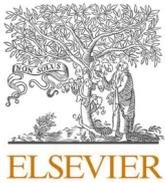
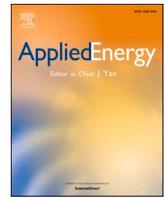

# Energy management of multi-mode plug-in hybrid electric vehicle using multi-agent deep reinforcement learning☆

Min Hua [a], Cetengfei Zhang [a], Fanggang Zhang [a], Zhi Li [b], Xiaoli Yu [b], Hongming Xu [a], Quan Zhou [a,b,*]

[a] *School of Engineering, University of Birmingham, Birmingham B15 2TT, UK*
[b] *State Key Laboratory of Clean Energy Utilization, Zhejiang University, Hangzhou 310027, China*

## HIGHLIGHTS

- A MADRL framework is proposed for multi-mode PHEV control.
- A relevance ratio is introduced for hand-shaking learning in the MADRL framework.
- The sensitivity of the impact factors for learning performance is ranked.
- A comparison between the single-agent and multi-agent systems is conducted.

## ARTICLE INFO

## ABSTRACT

The recently emerging multi-mode plug-in hybrid electric vehicle (PHEV) technology is one of the pathways making contributions to decarbonization, and its energy management requires multiple-input and multiple-output (MIMO) control. At the present, the existing methods usually decouple the MIMO control into single-output (MISO) control and can only achieve its local optimal performance. To optimize the multi-mode vehicle globally, this paper studies a MIMO control method for energy management of the multi-mode PHEV based on multi-agent deep reinforcement learning (MADRL). By introducing a relevance ratio, a hand-shaking strategy is proposed to enable two learning agents to work collaboratively under the MADRL framework using the deep deterministic policy gradient (DDPG) algorithm. Unified settings for the DDPG agents are obtained through a sensitivity analysis of the influencing factors to the learning performance. The optimal working mode for the hand-shaking strategy is attained through a parametric study on the relevance ratio. The advantage of the proposed energy management method is demonstrated on a software-in-the-loop testing platform. The result of the study indicates that the learning rate of the DDPG agents is the greatest influencing factor for learning performance. Using the unified DDPG settings and a relevance ratio of 0.2, the proposed MADRL system can save up to 4% energy compared to the single-agent learning system and up to 23.54% energy compared to the conventional rule-based system.

## 1. Introduction

Demands for advancement in veihicle performance and decarbonization motivate the automotive industry to move towards automation and electrification based on intelligent optimization [1–3]. Electrified vehicles, including plug-in hybrids, battery electric, and fuel cell vehicles are the keys to road transport electrification [4]. The energy management system (EMS) is a critical function module for electrified vehicles, which should be dedicated to various powertrain architectures and thus capable of maximizing energy efficiency while maintaining the health of the powertrain components [5]. The recently developed multi-mode PHEV utilizes a new powertrain topology to allow the vehicle to operate in pure battery mode, series hybrid mode, or parallel hybrid mode adaptively according to the driving conditions [6]. This powertrain topology has been adopted by some OEMs and T1 suppliers worldwide, e.g., Honda, BYD, and MAHLE [7]. Different from series or parallel HEVs, the multi-mode vehicle cannot use the coupled control of the engine, generator, and traction motor, and thus MIMO control is required.

There are currently three main categories of control strategies for the EMS, i.e., rule-based methods, optimization-based methods, and learning-based methods [8–10]. The rule-based control strategies typically implement deterministic rules or fuzzy logic that are founded on the parameters of the vehicle and expert knowledge, and they are computationally efficient and easy to be applied in real-time [11]. The optimization-based strategies include the model predictive control (MPC) [12], dynamic programming (DP) [13], Pontryagin's minimum principle (PMP) [14], and equivalent consumption minimization strategy (ECMS) [15]. They provide access to optimize vehicle performance in certain conditions. The main drawback of optimization-based strategies is that they have limited adaptability to real-world conditions, especially for dramatically changing conditions. It is tough to obtain good results for multi-objective and multi-mode optimization problems since they require heavy-duty computation to resolve the control models [16].

The learning-based EMSs are emerging in recent years and demonstrated their advantages in optimizing control policies during real-world driving [17,18]. Q-learning, a prevalent RL method at cost of computation and algorithmic complexity, has been developed for the EMSs with discretized state and action spaces. Qi et al. utilized the Q-learning algorithm to optimize the EMS for charging-depletion conditions [19]. Liu et al. proposed a Q-learning-based EMS by combining neurodynamic programming with future trip information under a two-stage deployment [20]. Chen et al. formulated a new EMS by incorporating the Q-learning algorithm with a stochastic model predictive control (SMPC), where a Markov chain-based velocity prediction model is developed to achieve superior fuel economy [21]. Zhou et al. proposed a multi-step Q-learning algorithm to enable the all-life-long online optimization of a model-free predictive EMS control [22]. Shuai et al. developed a double Q-learning algorithm for the hybrid vehicle by proposing two new heuristic action execution policies, the max-value-based policy and the random policy [23]. However, with the increasing number of state and action variables in advanced decision-making tasks, it is more difficult for Q-learning algorithms to compute all Q values corresponding to the discrete state-action pairs from the perspectives of computing efficiency and optimality. To overcome this shortcoming of Q-learning algorithms, deep reinforcement learning (DRL) algorithms are used for high-dimensional continuous decision-making tasks by using neural networks to approximate the value function outputs [24–27].

DRL algorithms have been studied by many researchers to deal with multiple-input and single-output (MISO) control in series hybrids, parallel hybrids, and power-split hybrids, which normally involve high-dimension input spaces, such as the SoC, power demand, vehicle velocity, etc. Some of the states need to be obtained through multi-information fusion by combining inverse smoothing and grey prediction fusion to mitigate the impact of sensor measurement errors caused by the sample rate discrepancy and time delay between multi-sensors [28]. Xiong et al. presented a DRL-based EMS for the PHEV, where the power transition probability matrices were attained from different new driving cycles with different Kullback-Leibler (KL) divergence rates [29]. Tang et al. proposed a DRL-based EMS method by incorporating two distributed DRL models including an asynchronous advantage actor-critic (A3C) model and a distributed proximal policy optimization (DPPO) model. The results demonstrated that the distributed DRL had laid a very good algorithm foundation for future work [30]. Zou et al. developed an accelerated DRL method by prioritizing a replay module in the deep Q network and an online-updated strategy for a fix-line hybrid electric vehicle [31]. Since they have provided much better control performance compared to the Q-learning methods and rule-based methods, DRL-based control strategies combined with other advanced algorithms have been employed in the EMS of many single-mode hybrid vehicles [32,33].

Studies of RL-based control for the multi-mode PHEV, however, are just at the beginning, because the multi-mode PHEV itself is new [34–36]. Tang et al. proposed a DRL method for the control of a multi-mode HEV, in which the deep Q-network (DQN) algorithm is used for gear shifting and the DDPG algorithm is used for controlling engine throttle [37]. Sun et al. developed a hierarchical power-splitting strategy that implements two DRL agents for multi-mode PHEV [38]. Jendoubi et al. incorporated hierarchical RL and multi-agent reinforcement learning (MARL) framework for efficient energy management and communication-free control to address multi-dimensional, multi-objective power system problems [39]. Shi et al. proposed a simple and robust EMS based on independent Q-learning (IQL) to achieve multi-stack fuel cell systems control of HEV by maintaining battery state of charge (SOC) and minimizing hydrogen consumption [40]. Wang et al. designed a MARL-based energy-saving strategy for hybrid electric vehicles (HEV) with advanced cruise control systems by combining powertrain and car-following behaviors to minimize energy consumption while maintaining a safe following distance [41]. To enable MIMO control with conventional RL/DRL algorithms that are only capable of MISO control, the above-mentioned methods implemented more than two RL/DRL agents to obtain the control outputs, but these RL/DRL agents have no links or communications and thus can only achieve local optimal results. Because the above-mentioned RL/DRL algorithms can only deal with single output control and are thus not capable of global optimization, the main objective of this paper is to develop a new type of RL algorithm that has multiple agents working with strong links for MIMO control of the multi-mode PHEV.

Multi-agent deep reinforcement learning (MADRL) is a recent breakthrough in artificial intelligence emphasizing the behaviors of multiple learning agents coexisting in a shared environment [42]. It links multiple RL agents in three working modes: 1) cooperative mode, 2) competitive mode, and 3) a mixture of the two [43]. In cooperative scenarios, agents work together to maximize a shared long-term return; in contrast, in competitive scenarios, agents' returns typically add up to zero; in mixed scenarios, there are general sum returns in both cooperative and competitive agents.

So far, MADRL has been explored in some areas, such as games, smart grid, and robots. However, it has never been used for the EMS of PHEV. The authors of this paper believe that MARL is a good solution to the MIMO control in the multi-mode PHEV. Therefore, the presented work has been focused on developing such EMS based on MADRL with three new contributions: 1) the best setting for the DDPG agents has been obtained through a parametric study, concerning network layers, learning rate, and policy noise; 2) a hand-shaking strategy has been developed by introducing a relevance ratio that facilitates collaboration among the DDPG agents and synchronize their learning processes towards the global goal; 3) multiple optimization objectives, including minimizing fuel consumption and minimizing battery SoC sustaining error, have been studied by implementing the proposed multi-agent

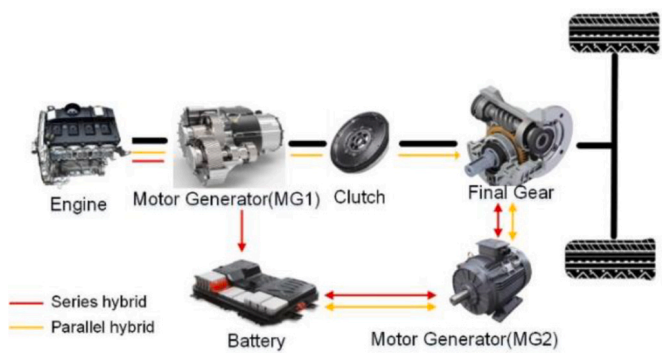

**Fig. 1.** Configuration of the multi-mode PHEV powertrain.

learning system in a multi-mode PHEV.

The rest of this paper is organized as follows: Section 2 formulates the MIMO control problem by modeling the multi-mode PHEV system. The MADRL framework is proposed in Section 3, with a DDPG-based EMS introduced as the baseline method. Test and validation are conducted on a software-in-the-loop platform and the results are discussed in Section 4. Section 5 summarizes the conclusions.

## 2. MIMO control of a multi-mode PHEV

The architecture of the multi-mode PHEV studied in this paper is shown in Fig. 1. The motor generator (MG1) and the engine work together to maintain the battery's SoC at a certain level for safety. The other motor (MG2) and engine are the power sources for driving. The multi-mode PHEV can work in different modes, which are controlled by engaging or disengaging the clutch, as illustrated in Fig. 1. In series mode (red lines), the clutch is disengaged and only MG2 drives the powertrain. In parallel mode (yellow lines), the clutch is engaged, and the engine will provide the part of driving torque as the supplement to the MG2.

### 2.1. The energy flow model

The energy flow of the vehicle is modeled based on longitudinal vehicle dynamics, and the force demand, $F_{dem}(t)$, the power demand, $P_{dem}(t)$, and the torque demand, $T_{dem}(t)$, can be calculated by

$$F_{dem}(t) = mgf\cos\alpha + \frac{1}{2}\rho A_f C_d v^2(t) + mg\sin\alpha + ma \tag{1}$$

$$P_{dem}(t) = F_{dem}(t) \bullet v \tag{2}$$

$$T_{dem}(t) = F_{dem}(t) \bullet R \tag{3}$$

where $m$ is the vehicle mass; $a$ is the vehicle acceleration, $g$ is the gravity acceleration; $f$ is the rolling resistance coefficient; $\rho$ is the air density; $A_f$ is the front area of the vehicle; $C_d$ is the air resistance coefficient; $v$ is the longitudinal velocity; $R$ is the wheel radius; $\alpha$ is the road slope, in this paper, the road slope should not be considered. Note that vehicle velocity and acceleration can be estimated accurately and reliably by synthesizing the vehicle dynamics and kinematics in a consensus framework [44].

In the series mode, the energy flow is described as:

$$\left. \begin{array}{l} P_{dem}(t) = P_{mot2}(t) \\ T_{dem}(t) = i_2 \bullet T_{mot2}(t) \\ P_{eng}(t) = \dfrac{n_{eng}(t) \bullet T_{eng}(t)}{9550} \\ P_{eng}(t) = P_{mot1}(t) \\ n_{eng}(t) = n_{mot1}(t) \end{array} \right\} \tag{4}$$

In the parallel mode, the energy flow is described as:

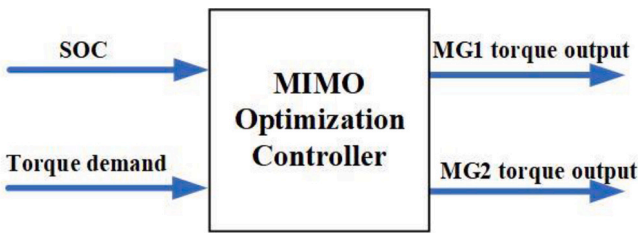

**Fig. 2.** MIMO control architecture.

$$\left. \begin{array}{l} P_{dem}(t) = P_{mot2}(t) + \left(P_{eng}(t) - P_{mot1}(t)\right) \\ T_{dem}(t) = i_1 \bullet \left(T_{eng}(t) - T_{mot1}(t)\right) + i_2 \bullet T_{mot2}(t) \\ P_{eng}(t) = \dfrac{n_{eng}(t) \bullet T_{eng}(t)}{9550} \\ n_{eng}(t) = n_{mot1}(t) \end{array} \right\} \tag{5}$$

where, $P_{mot1,2}(t)$ and $P_{eng}(t)$ are the electric power of MG1 and MG2, respectively; $n_{mot1}(t)$, $n_{mot2}(t)$, and $n_{eng}(t)$ are the rotational speed of MG1, MG2, and the engine, respectively; the MG1 and MG2 torque are separately $T_{mot1,2}(t)$ and $T_{eng}(t)$ is the engine torque; $i_1$ is the transmission ratio of the gearbox to MG1, $i_2$ is the final ratio of the gearbox from MG2 to the wheels.

The energy of a multi-mode PHEV is from the battery and the engine (fuel tank), and the total power loss mainly consists of the engine loss, $Loss_{eng}(t)$, and battery loss, $Loss_{batt}(t)$, which can be calculated by:

$$\left. \begin{array}{l} P_{loss}(t) = Loss_{eng}(t) + Loss_{batt}(t) \\ Loss_{eng}(t) = \dot{m}_f(t) \bullet H_f - \dfrac{n_{eng}(t) \bullet T_{eng}(t)}{9550} \\ Loss_{batt}(t) = R \bullet I_{batt}(t)^2 \end{array} \right\} \tag{6}$$

where, $P_{loss}$ is the total power loss, $H_f$ is the heat value of fuel ($H_f = 43.5\ KJ/g$); and $R$ is the equivalent internal resistance in the battery model.

a) **Engine model**

The engine model is used to determine the fuel consumption rate $\dot{m}_f$ (g/s) based on a 2D look-up table, which is a function of the engine speed, $n_{eng}(t)$, and the engine torque, $T_{eng}(t)$, as

$$\dot{m}_f(t) = f\left(n_{eng}(t), T_{eng}(t)\right) \tag{7}$$

b) **Motor models**

The power demands of MG1 and MG2 are modeled based on two quasi-static energy efficiency maps $\eta_1$ and $\eta_2$, respectively, as follows:

$$P_{mot1}(t) = \dfrac{n_{mot1}(t) \bullet T_{mot1}(t)}{9550} \bullet \eta_1\left(n_{mot1}(t), T_{mot1}(t)\right) \tag{8}$$

$$P_{mot2}(t) = \begin{cases} \dfrac{n_{mot2}(t) \bullet T_{mot2}(t)}{9550} \bullet \eta_2\left(n_{mot2}(t), T_{mot2}(t)\right); T_{mot2}(t) \leq 0 \\ \dfrac{n_{mot2}(t) \bullet T_{mot2}(t)}{9550} \bullet \dfrac{1}{\eta_2\left(n_{mot2}(t), T_{mot2}(t)\right)}; T_{mot2}(t) > 0 \end{cases} \tag{9}$$

c) **Battery model**

The battery model is established based on an equivalent circuit as follows:

$$\left. \begin{array}{l} P_{batt}(t) = P_{mot2}(t) - P_{mot1}(t) \\ P_{batt}(t) = U \bullet I_{batt}(t) - R \bullet I_{batt}^2(t) \\ I_{batt}(t) = \dfrac{U - \sqrt{U^2 - 4 \bullet R \bullet P_{batt}(t)}}{2 \bullet R} \\ SoC(t) = SoC(0) - \dfrac{\int_0^t I_{batt}(t)dt}{Q_{batt}} \end{array} \right\} \tag{10}$$

where, $P_{batt}(t)$ is the output power of the battery pack during charging and discharging; $SoC(0)$ is the initial SoC value; $I_{batt}(t)$ is the current of the battery at time t; and $Q_{batt}$ is the nominal battery capacity ($Q_{batt}$ =54.3 Ah). Since this research focuses on the energy flow in different driving cycles (up to 1800 s), the battery temperature and aging dynamics are ignored. The open-circuit voltage (OCV) $U$ ($U = 350$ V) and the battery internal resistance $R$ ($R = 0.15$ Ω) are both constant.

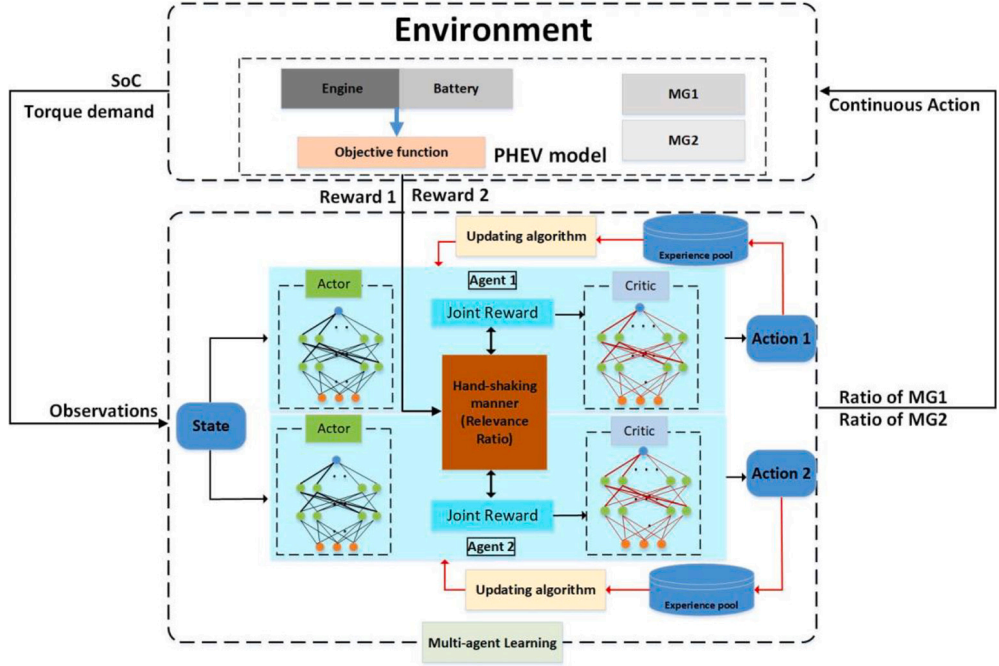

**Fig. 3.** DDPG-based EMS with multi-agent learning.

*2.2. The multiple-input and multiple-output (MIMO) energy management controller*

A multiple-input and multiple-output (MIMO) controller, as shown in Fig. 2, is developed to manage the energy flow of the studied vehicle. By observing the battery SoC and the overall torque demand as the control inputs, the MIMO controller calculates the torque demands for MG1, MG2, and the engine, respectively, to provide sufficient general torque to drive the vehicle while maintaining the battery SoC. The proposed energy management method is built on the assumption that the variability and uncertainty of control inputs can be ignored, implying that the input data is assumed to be accurate and fully observable.

The core of the MIMO control is to resolve an optimization problem defined as follows:

$Minimize\ P_{loss}(u_{mot1},u_{mot2},T_{dem})\ and\ \Delta SOC(u_{mot1},u_{mot2},T_{dem})$

$$s.t. \begin{cases} Loss_{eng}(t) = \dot{m}_f(t) \bullet H_f - \dfrac{n_{eng}(t) \bullet T_{eng}(t)}{9550} \\ Loss_{batt}(t) = R \bullet I_{batt}(t)^2 \\ SoC(t) = SoC(0) - \dfrac{\int_0^t I_{batt}(t)dt}{Q_{batt}} \\ SoC^- \leq SoC(t) \leq SoC^+ \\ and\ other\ physical\ constraints \end{cases} \quad (11)$$

where the overall power loss, $P_{loss}$, and the SoC difference, $\Delta SOC$, are two objectives that need to be minimized; the MG1 torque command, $u_{mot1}(t)$, and the MG2 torque command, $u_{mot2}(t)$, are the optimization variables to be determined during the real-time control. The optimization should be subjected to the vehicle energy flow models

**Table 1**
Comparison of single-agent and the proposed multi-agent systems.

| | Single-agent | The proposed multi-agent |
|---|---|---|
| No. of Agents | 1 | 2 |
| States | $T_{dem}(t), SoC(t)$ | $T_{dem}(t), SoC(t)$ |
| Action(s) | $u_{mot1}(t)$ | $u_{mot1}(t), u_{mot2}(t)$ |
| Reward | Weighted sum | Two unique functions for different preferences |

and other physical constraints of the powertrain system and subsystems.

## 3. Hand-shaking multi-agent learning for MIMO control

*3.1. Multi-agent learning with the DDPG algorithm*

To resolve the optimization problem defined in Eq. (11), this paper proposes a hand-shaking multi-agent learning scheme, as shown in Fig. 3, in which two DDPG agents are involved to minimize the fuel consumption and battery usage simultaneously through the torque control of MG1 and MG2. Each learning agent has an actor-network and a critic-network. In each time interval, the agent starts learning with an observation of the state variables, and it uses the actor-network to generate a control action for the vehicle system followed by a reward evaluation from the system feedback. The critic-network is trained to update the actor-network using a policy gradient algorithm based on the recorded variables of state, action, and reward. It is worth noting that the two rewards are obtained through a hand-shaking manner module, where they are considered as two joint rewards. Further details regarding these rewards will be provided later.

The differences between the multi-agent system and the conventional single-agent system (baseline) are summarized in Table 1. Details of the main components of both learning systems are described as follows. The main difference between the single-agent system and the multi-agent system is the number of learning agents, i.e., the single-agent system only has one learning agent while the multi-agent system has more than two agents. The learning agent is a multi-input and single-output (MISO) control model that has the capability of self-learning for the development of control policy. It can be developed based on Q-learning, deep Q-learning, or other RL algorithms. In this study, the environment states and action variables are continuously varying. Therefore, the DDPG agent has been developed. The proposed multi-agent system includes two agents with different reward preferences for the two optimization objectives, and each agent generates the control signal for MG1 and MG2, respectively.

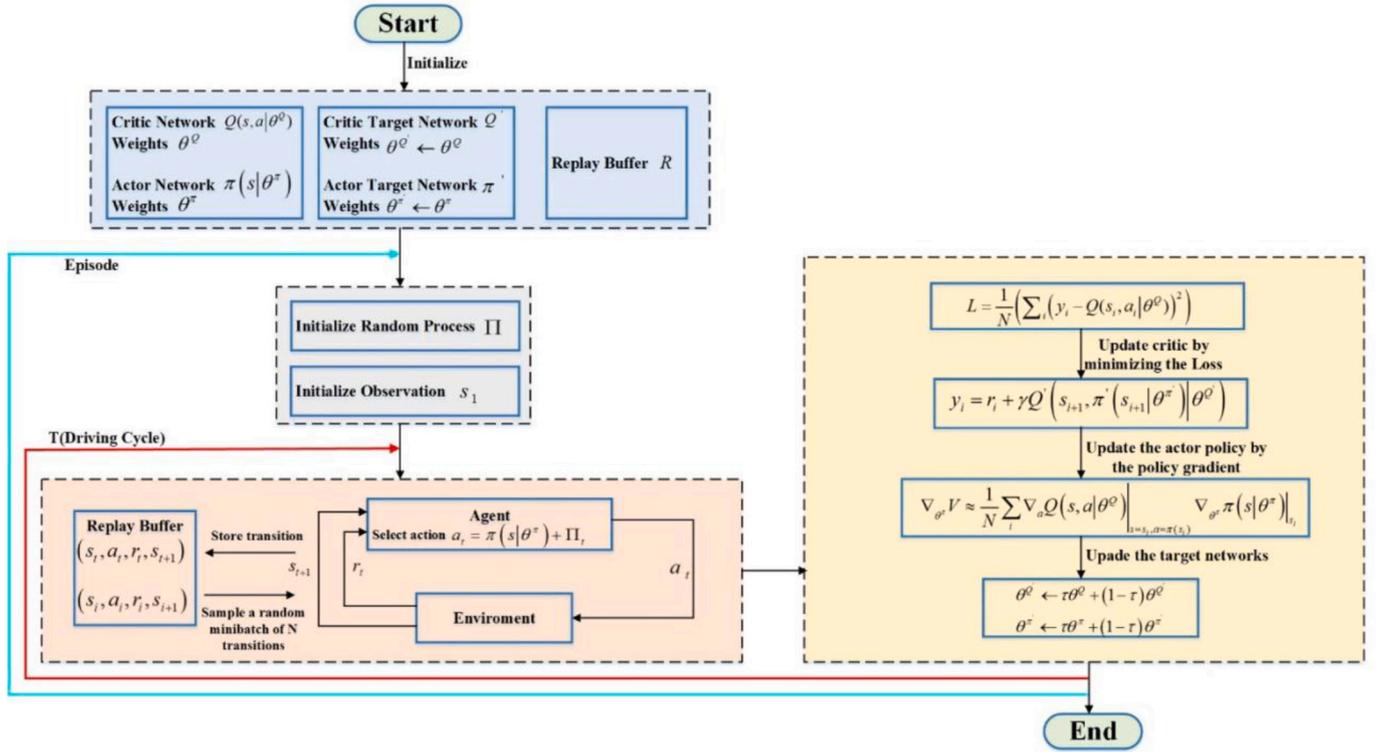

**Fig. 4.** Control policy optimization using the DDPG algorithm

### 3.2. DDPG-based learning process

This paper implements the deep deterministic policy gradient algorithm to optimize the control policy with an outer loop and an inner loop as shown in Fig. 4. In the inner loop shown with the red line, the agent interacts with the vehicle driving in real-world in with a sampling time of 1 s. Once the vehicle finished driving within a certain time defined in

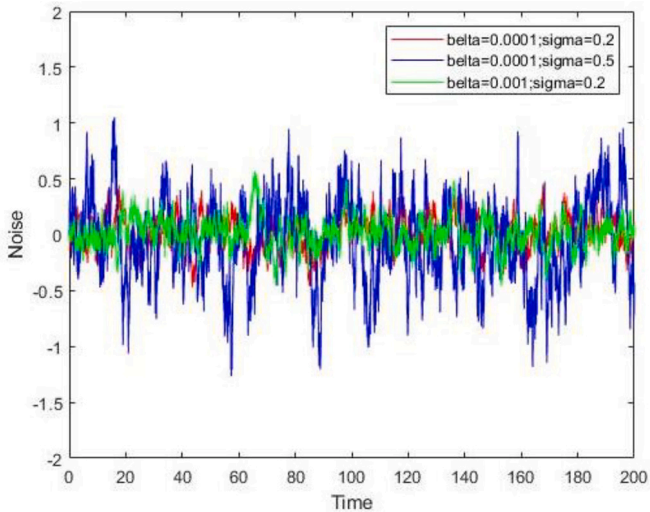

**Fig. 5.** Ornstein-Uhlenbeck (OU) noise process under different parameters.

a driving cycle, the agent will update the policy in the outer loop as illustrated with the cyan line [37]. (See Fig. 4)

By defining a policy, $\pi(s)$, as

$$a = \pi(s) = argmax Q(s,a) \tag{12}$$

where $a$ is the control action; and $s$ is the state variables. If the policy, $\pi : S \leftarrow A$, is deterministic with the actor-network, the policy update process can be formulated as statistical learning of the critic-network, $Q(s,a)$, by

$$Q^\pi(s_t,a_t) \leftarrow E_{r_t,s_{t+1}\sim E}[r(s_t,a_t) + \gamma Q^\pi(s_{t+1},\pi(s_{t+1}))] \tag{13}$$

where $r$ is the reward variable; and $\gamma$ is the discount factor. In the statistical learning process, an experience replay buffer with size $3R$ is used to store the transitions, which is a time-series batch of states, actions, and rewards in a form like $s_t,a_t,r_t,s_{t+1},a_{t+1},r_{t+1},\ldots a_{t+R},r_{t+R}$. Since DDPG is an off-policy algorithm, the buffer size can be very big to allow the system to learn from a large number of unrelated transitions. To reduce the computation load, this paper implements a minibatch method that randomly selects $N$ samples ($N \ll R$) from the experience buffer to train the actor and critic networks at different times.

Once a new batch of data is collected from real-world driving, this paper implements the temporal difference (TD) method to estimate the hyperparameters of the critic network, $\theta^Q$, by minimizing a loss function, $L$, defined by

$$L(\theta^Q) = E_{s_t\sim\rho, a_t\sim\pi, r_t\sim E}\left[\left(Q(s_t,a_t|\theta^Q) - r(s_t,a_t) + \gamma Q(s_{t+1},\pi(s_{t+1})|\theta^Q)\right)^2\right] \tag{14}$$

Then, by implementing a Deterministic Policy Gradient (DPG) method, the actor network can be updated by

$$\nabla_{\theta^\pi} V \approx \frac{1}{N}\sum_i \nabla_a Q(s,a|\theta^Q)\Big|_{s=s_i,a=\pi(s_i)} \nabla_{\theta^\pi}\pi(s|\theta^\pi)\Big|_{s=s_i} \tag{15}$$

Since the critic network $Q(s,a|\theta^Q)$ is updated and used to estimate the target value, the Q update is prone to be unstable in many environments. Target networks, $Q'(s,a|\theta^{Q'})$ and $\pi'(s|\theta^{\pi'})$, are employed for the actor and critic networks respectively to calculate the target values to provides momentum to the learning process by introducing a factor $\tau \ll 1$ to weights the parameters by $\theta' \leftarrow \tau\theta + (1-\tau)\theta'$.

Exploration is one of the most challenging problems of the learning process in continuous action domains. An exploration policy $\pi'$ by combining the actor policy with a noise process $\Pi$, which is described as:

$$\pi'(s_t) = \pi'(s_t|\theta_t^{\pi}) + \Pi \tag{16}$$

An Ornstein-Uhlenbeck (OU) process [40] is chosen to produce the noise by:

$$da_t = -\beta a_t dt + \sigma dW_t \tag{17}$$

where $W_t$ is a Wiener process with normally distributed increments, the decay rate $\beta > 0$ (how "strongly" the system reacts to perturbations) and the variation $\sigma > 0$ of the noise should be set and tunned, as shown in Fig. 5. Since the OU process is time-series related and can be used to generate temporally correlated exploration in the action selection process of the previous step and the next step of RL to improve the exploration efficiency of control systems.

### 3.3. States and actions

In this study, both the single-agent system and the multi-agent system monitor vehicle torque demands and battery SoC values as the state variables in a two-dimensional vector space, $s(t)$, as

$$s(t) = [T_{dem}(t), SoC(t)] \tag{18}$$

where $T_{dem}(t)$ is the power demand of the vehicle and $SoC(t)$ is the battery state-of-charge level.

As shown in Table 1, the single agent system can only output a single control action, $a_s(t)$, the control command of MG2 $u_{mot2}(t)$ keep the constant, and this study uses the DDPG algorithm to compute the control command of MG1, $u_{mot1}(t)$, based on a deterministic policy $\mu_{sa}$:

$$a_s(t) = u_{mot1}(t) = \pi_{sa}(s(t)|\theta^{\pi_{sa}}) \tag{19}$$

where, $\theta_{sa}(t)$ is a hyper-parameter matrix representing the control policy that is updated over time. And the control commands of MG1 and MG2, i.e., $u_{mot1}(t)$ and $u_{mot2}(t)$, were dynamically optimized through multi-agent learning. Based on $u_{mot1}(t)$ and $u_{mot2}(t)$, the control command $u_{eng}(t)$ of the engine can be calculated by

$$T_{eng} = u_{mot1}(t) \bullet T_{mot1\_max} + T_{GB}$$
$$T_{dem} = u_{mot2}(t) \bullet T_{mot2\_max} + T_{eng} \tag{20}$$
$$u_{eng}(t) = \frac{u_{mot1}(t) \bullet T_{mot1\_max} + T_{GB}}{T_{eng\_max}}$$

where, $T_{mot1\_max}$ is the maximum torque of the MG1; $T_{eng\_max}$ is the maximum torque that can be supplied by the engine, and $T_{dem}$ is the torque demand for driving and braking the vehicle; $T_{GB}$ is the torque of the output shaft provided by the engine when MG2 output torque cannot meet the requirement of the total torque output.

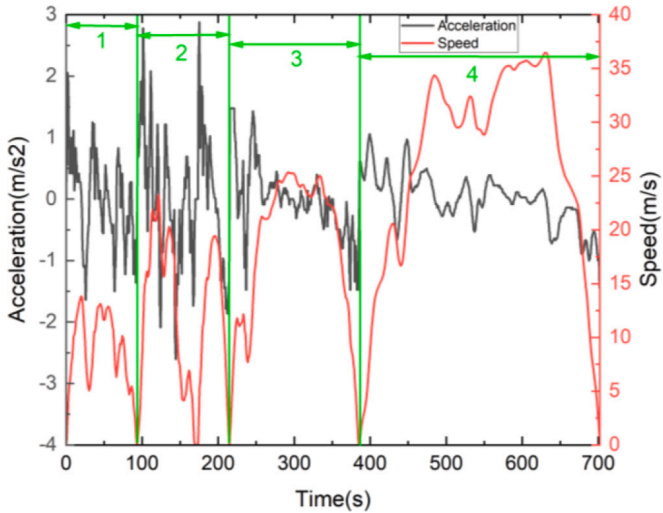

**Fig. 6.** An example of the learning cycle.

For the proposed multi-agent system that has two DDPG agents, the actions as the output of the multi-agent system can be expressed:

$$\begin{cases} a_m(t) = [u_{mot1}(t), u_{mot2}(t)] \\ u_{mot1}(t) = \pi_{ma1}(s(t)|\theta^{\pi_{ma1}}) \\ u_{mot2}(t) = \pi_{ma2}(s(t)|\theta^{\pi_{ma2}}) \end{cases} \tag{21}$$

where $u_{mot1}(t)$ is the output of the first DDPG agent for control of MG1 while $u_{mot2}(t)$ is the output of the second DDPG agent for MG2. And the engine control command can be calculated using Eq.20. Both $a_s(t)$ and $a_m(t)$ are calculated following a rolling process of exploration and exploitation [18].

### 3.4. Reward functions and hand-shaking design

The single-agent system implements a weighted sum method to incorporate the optimization objectives by

$$r_s(t) = -\alpha P_{loss}(t) - \beta|SoC_{ref} - SoC(t)| \tag{22}$$

where $\alpha$ is a scaling factor; $SoC_{ref}$ is the target battery SoC value to be maintained during the driving; and $\beta$ is a conditional weight factor, which yields:

$$\beta = \begin{cases} 0, SoC(t) \geq SoC_{ref} \\ 2, SoC(t) < SoC_{ref} \end{cases} \tag{23}$$

The conditional weight factor, $\beta$, will allow the DDPG agent to have a higher priority in minimizing fuel consumption when the SoC level is high [32].

A hand-shaking strategy is developed for the multi-agent system by introducing a relevance ratio $R_{rel}$ that facilitates collaboration among the DDPG agents in the MADRL system. It ensures that the agents synchronize their learning processes and maintain stability during the training process. The relevance ratio aims to allow the agents to exchange information effectively through joint rewards for optimal control decisions. It incorporates global reward ($r_{global}$), and local rewards ($r_{local,1}$ and $r_{local,2}$) by

$$\begin{cases} r_{m1} = R_{rel} * r_{global} + r_{local,1} \\ r_{m2} = R_{rel} * r_{global} + r_{local,2} \end{cases} \tag{24}$$

where $r_{m1}$ and $r_{m2}$ are the joint rewards for the first DDPG agent and the second DDPG agent, respectively. Since minimizing the power loss is the main optimization objective, the power loss value is the element for the global reward function,

$$r_{global} = -P_{loss}(t) \tag{25}$$

Two local reward functions are designed to balance the usages of the ICE engine and the battery with two DDPG agents. $r_{local,1}$ and $r_{local,2}$ are allocated for the first DDPG agent and the second DDPG agent, respectively, and they can be calculated by:

$$\begin{cases} r_{local,1} = -\beta|SoC_{ref} - SoC(t)| \\ r_{local,2} = -\alpha Loss_{eng}(t) \end{cases} \tag{26}$$

where $\alpha$ is the scaling factor, and $\beta$ is the weighting factor. In this research, $\alpha$, and $\beta$ in the multi-agent system are set the same as in the single-agent system.

**Table 2**
Main hyperparameters and their values.

| Hyperparameter | Values |
| --- | --- |
| Discount factor | 0.99 |
| Batch size | 64 |
| Experience buffer | $1 \times 10^5$ |
| Regulation factor for both networks | $1 \times 10^{-4}$ |
| Policy noise with a decay rate | 0.2, 0.5 |
| Policy noise with a variation | $1 \times 10^{-3}, 1 \times 10^{-4}$ |
| Optimizer | Adam |
| Critic networks layers | 2–7 |
| Actor networks layers | 3 |
| Learning rate | $1 \times 10^{-4}, 1 \times 1^{-3}, 1 \times 10^{-5}$ |

## 4. Results and discussion

A software-in-the-loop (SiL) testing platform is built for testing and validation on a workstation with an i7-7600U CPU and 64GB RAM. The models including the vehicle plant model and MIMO control model for the SiL test are developed using MATLAB/Simulink version 2022a, where the solver was ODE1 and the sample time is 1 s. The impacts of the learning agent design on MIMO control performance are firstly studied with sensitivity ranked. A parametric study on the relevance ratio, $R_{rel}$, is conducted to attain the best handshaking strategy for multi-agent control optimization. The performance of the single-agent system and multi-agent system are compared on a multi-mode PHEV driving under a training cycle and two testing cycles. The training driving cycle is built with elements generated from four standard driving cycles, including

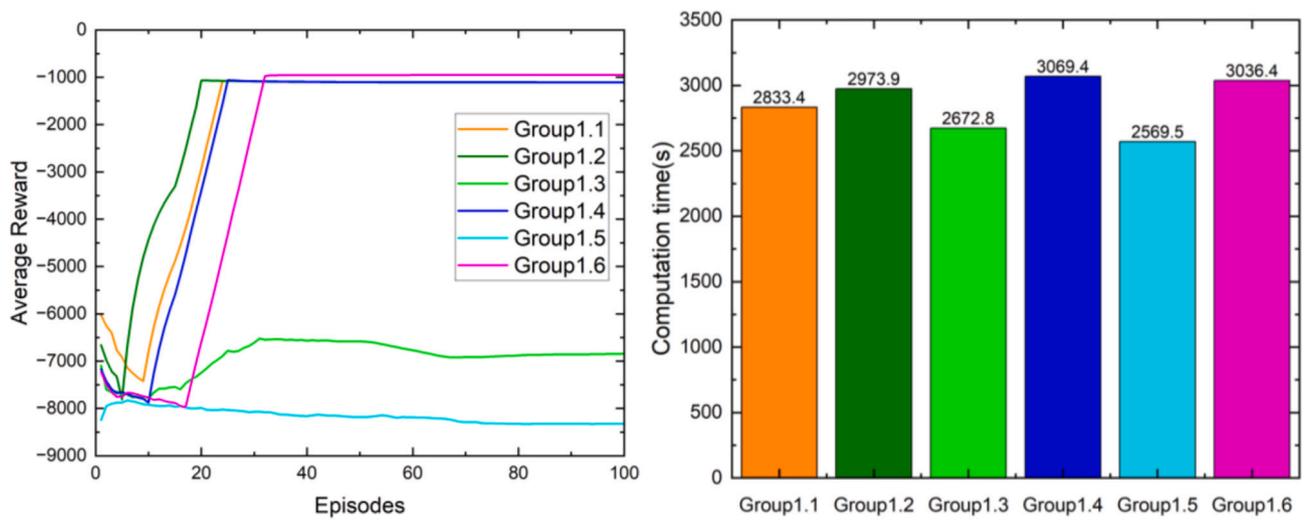

(a) The average rewards   (b) Computation time

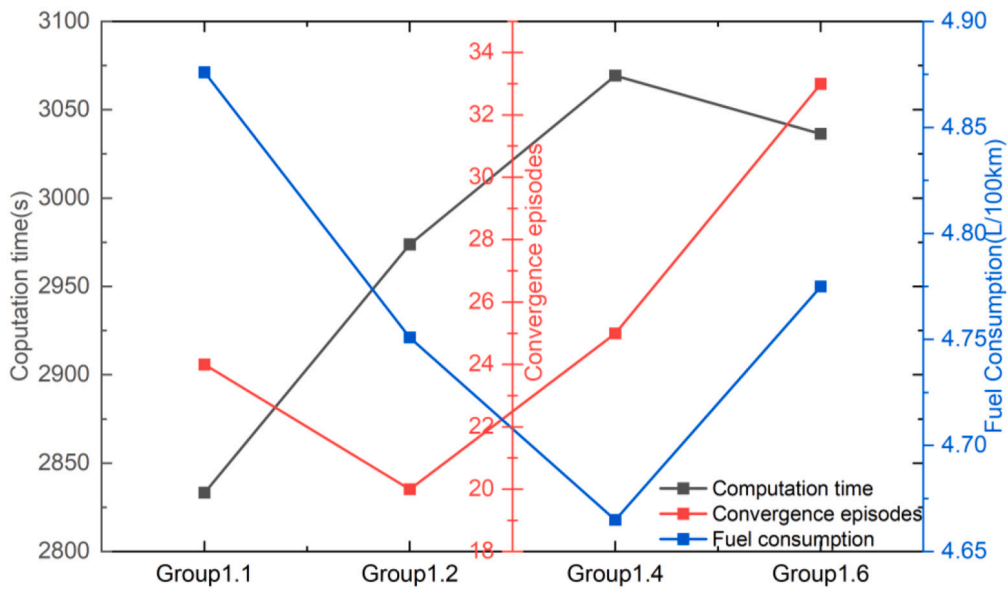

(c) Comparison with effective groups

**Fig. 7.** Impact of critic network layers on the control performance.

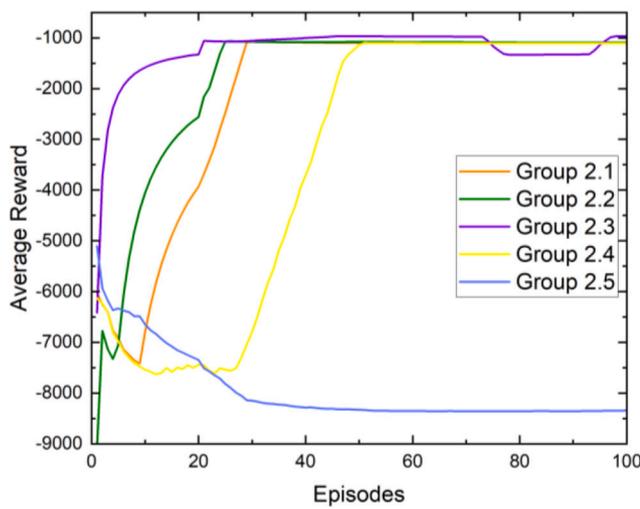

(a) The average rewards

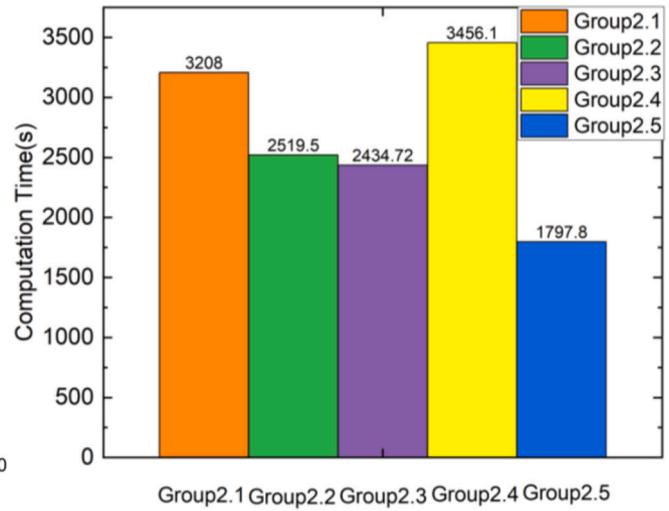

(b) Computation time

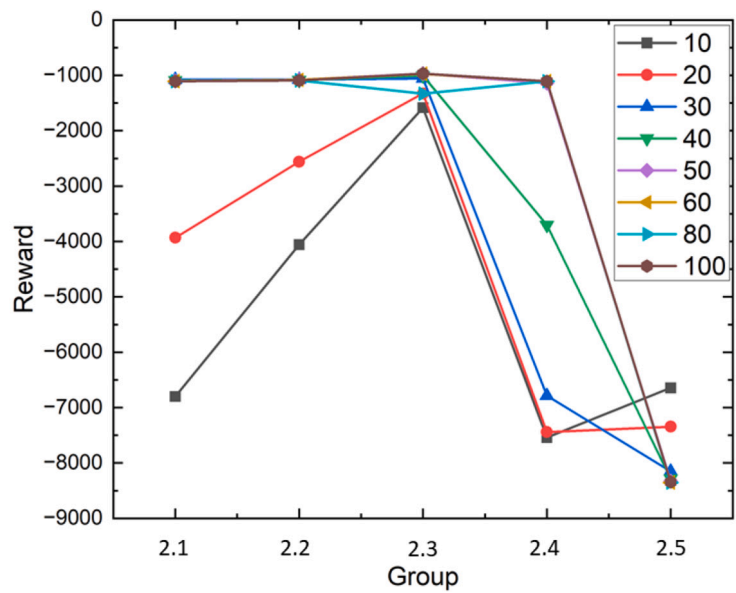

(c) Reward change curve under different episodes with 5 groups

**Fig. 8.** The results of different learning rates

Artemis Rural, RTS95, UDDS, and WLTP, as illustrated in Fig. 6. It has four phases, where Phase 1 represents the low-speed region of the Artemis Rural cycle; Phase 2 involves the maximum acceleration in the RTS95 cycle; Phase 3 represents the medium-speed in UDDS Driving Cycle; and Phase 4 involves the high-speed region in the WLTP Cycle. In each learning episode, the four phases will be reorganized randomly to provide the learning noise for the robustness evaluation of the learning.

### 4.1. Impacts of learning agent design on MIMO control performance

In this Section, the impacts of design parameters including the number of network layers, learning rate values, and different policy noise levels on the MIMO control performance are analyzed. The analysis is conducted based on the studied vehicle driving under the training cycle with an initial SoC value of 0.28. The main hyperparameters, which are vital to the learning dynamics and performances, are listed shown in Table 2.

a) *Networks layers*

Since the critic network is to obtain an accurate estimate of the Q function value for the evolution of the actor-network, this paper first focuses on the design of the critic network. We investigate the critic networks with different numbers of layers from 2 to 7 with an interval of 1 set for Group 1.1 to Group 1.6 while remaining the number of actor networks as a constant of 3. The tests are conducted with six individual groups to testify how the number of layers can affect MIMO control performance. The control performance including the average reward, the computer time, and the fuel economy are compared in Fig. 7.

Theoretically, the number of network layers is related to the capability of representing nonlinear models and affects the complexity of the trained objects. This is why the learning performance tends to be improved while the computing time increases. From the results shown in Fig. 7 (a) and (b), Group 1.3 and Group 1.5 perform much worse than others although their computation time is lower than others. Groups 1.1,

Table 3
Comparison of SOC and Fuel consumption.

| Group | 2.1 | 2.2 | 2.3 | 2.4 | 2.5 |
|---|---|---|---|---|---|
| Initial SOC | 0.28 | 0.28 | 0.28 | 0.28 | 0.28 |
| End SOC | 0.311 | 0.294 | 0.303 | 0.316 | – |
| Fuel economy (L/100 km) | 4.620 | 4.547 | 4.589 | 4.726 | – |

Table 4
Policy noise combination

| Group | 3.1 | 3.2 | 3.3 |
|---|---|---|---|
| $\beta$ | $1 \times 10^{-4}$ | $1 \times 10^{-4}$ | $1 \times 10^{-3}$ |
| $\sigma$ | 0.2 | 0.5 | 0.2 |

Table 5
Comparison of SOC and Fuel consumption

| Group | 3.1 | 3.2 | 3.3 |
|---|---|---|---|
| Initial SOC | 0.28 | 0.28 | 0.28 |
| End SOC | 0.294 | 0.313 | 0.309 |
| Fuel economy (L/100 km) | 4.547 | 4.689 | 4.625 |

Table 6
Comparison of importance levels

| | | Computation time (CT) | Convergence episodes (CE) | Fuel Economy (FE) |
|---|---|---|---|---|
| Networks layers (Group1) | The worst | 3069.40 | 33 | 4.876 |
| | The best | 2833.40 | 20 | 4.665 |
| | $L_s$ (%) | **7.90** | **59** | **1.430** |
| Learning rate (Group2) | The worst | 3456.10 | 50 | 4.726 |
| | The best | 2434.72 | 20 | 4.547 |
| | $L_s$ (%) | **33.124** | **118** | **6.186** |
| Policy noise (Group3) | The worst | – | 68 | 4.689 |
| | The best | – | 30 | 4.547 |
| | $L_s$ (%) | – | **89.566** | **1.397** |

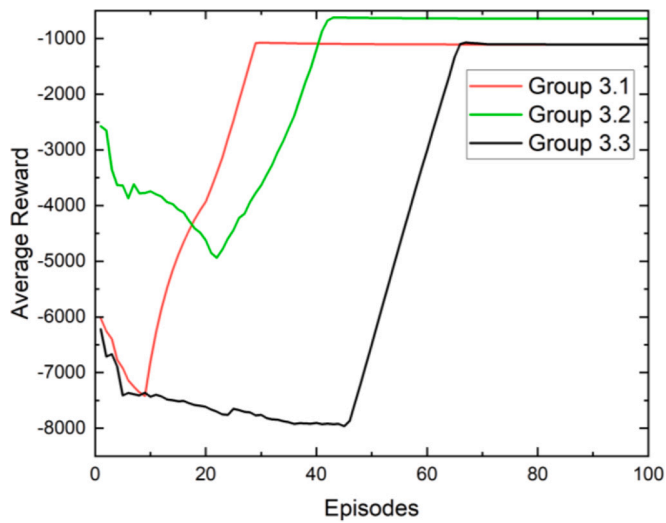

Fig. 9. The average rewards of different policy noise

1.2, 1.4, and 1.6 all received acceptable performance during the training process with different computation efforts. Group 1.6 increases performance less with more costs of time to achieve stability, the possible reason would be the overfitting of the critic network due to its deep architecture. According to a comprehensive comparison as illustrated in Fig. 7 (c), this paper suggests the critic network with parameters in Group 1.2 is the best concerning three aspects including convergence episodes, computation time, and fuel consumption.

b) *Learning rate*

Five groups of tests are designed to investigate the impact of the learning rates on the MIMO control performance. Groups 2.1, 2.2, and 2.5 have equal settings of learning rates for the actor and critic networks with values of $1 \times 10^{-4}$, $1 \times 10^{-3}$, and $1 \times 10^{-5}$, respectively. Groups 2.3 and 2.4 have different settings for the actor and critic networks. In Group 2.3, the actor learning rate is $1 \times 10^{-4}$, and the critic learning rate is $1 \times 10^{-3}$. In Group 2.4, the actor learning rate is $1 \times 10^{-3}$, and the critic learning rate is $1 \times 10^{-4}$. The other network settings are the same as Group 1.2. The learning and MIMO control performance are compared in Fig. 8 and Table 3.

From Fig. 8 (a) and (b), it is apparent the learning rate is crucial for MIMO control optimization and produces a distinguished impact on learning through the update of both networks. Theatrically, the learning agent with a higher learning rate achieves better learning performance, this is the reason why Group 2.2 outperforms Group 2.1 and 2.5. According to the result in Group 2.5, if the learning rate is too small, the learning process would be unstable. From the comparison of Groups 2.3 and 2.4, improving the learning rate of the critic network is shown more effective in developing the MIMO control performances. By investigating the learning performance at each episode from 0 to 100 with an interval of 10 episodes in Fig. 8 (c) and vehicle performance in Table 2, the MIMO controller with the settings in Group 2.3 is the best. It requires the least computation time and is the fastest to reach the coverage point, although its fuel consumption is 0.9% higher than Group 2.2.

c) *Policy noise*

As defined in Eq. 17, the OU process, dominating the ratio of exploration during reinforcement learning, is determined by the weighting value of the Wiener process, $\beta$, and the variation of the noise, $\sigma$. By implementing the network setting in Group 1.2 and Group 2.3, three groups of test with different values of $\beta$ and $\sigma$, as illustrated in Table 4 are conducted.

The three groups of policy noise combinations are implemented respectively in the studied vehicle driving under the learning cycle, and the average reward, battery SOC, and fuel economy (L/100 km) are compared in Fig. 9 and Table 5. The results indicate that the controller with the setting defined in Group 3.1 is the best since it requires less time to converge to the best system performance which leads to the best fuel economy and battery charge sustaining. The results also show that the weighting value of the Wiener process, $\beta$, impacts the learning speed more significantly compared to the variation of the OU process, $\sigma$.

d) *Importance analysis*

By introducing a sensitivity level, $R_s$, which is defined as

$$L_s = \frac{|y_{best} - y_{worst}| \bullet \|x_{best}\|_2}{y_{best} \bullet \|x_{best} - x_{worst}\|_2} \times 100\% \qquad (27)$$

where $y_{best}$ and $y_{worst}$ are the best-case values and the worse-case values of the indicators including computation time (CT), the number of episodes for convergence (CE), and fuel economy (FE); $x_{best}$ and $x_{worst}$ are projected values of the learning agent settings on a one-dimension vector for the best case and the worst case, respectively. The principal component analysis (PCA) method is utilized to project multi-dimension network setting matrices into the one-dimensional vector. The best-case values, the worst-case values, and the sensitivity levels of each group of tests are compared in Table 6. The results highlight that the learning rate is the most important influencing factor on learning performance in terms of CT, CE, and FE. Policy noise is the second contributor to the number of episodes for convergence (CE) while the number of network layers is the second contributor to fuel economy (FE). The policy noise has no contribution to computation time (CT).

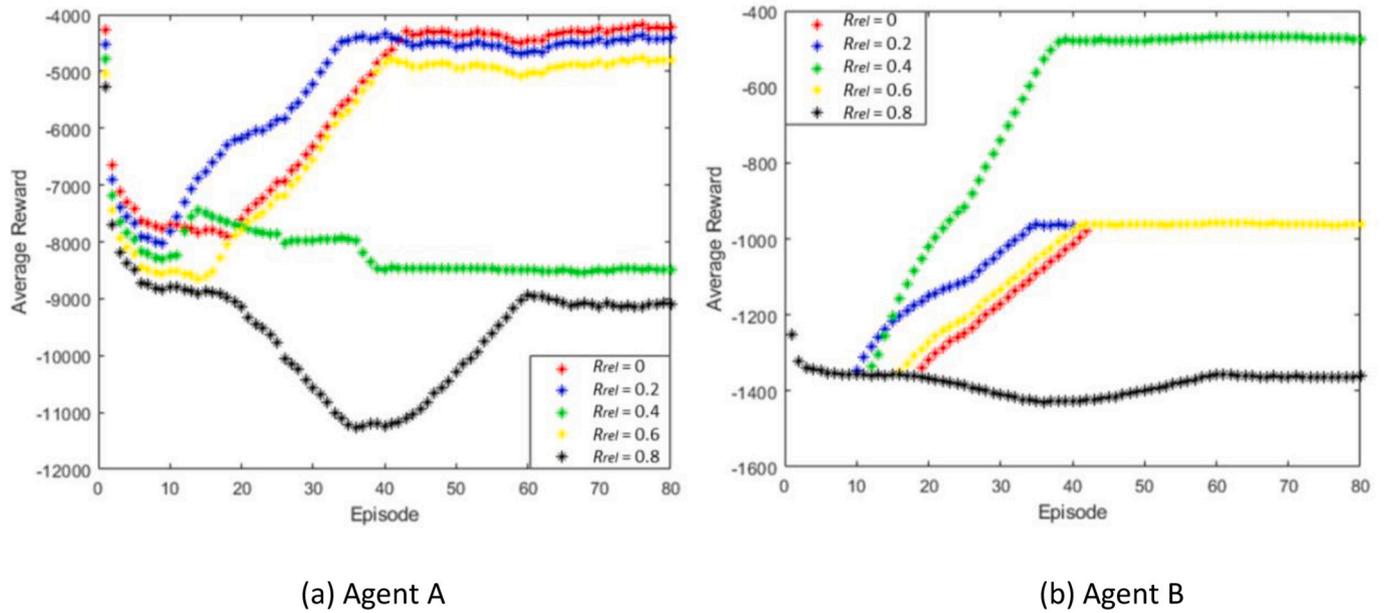

**Fig. 10.** Learning performance of different ratios (SOC initial value is 0.28).

Hand-shaking of the multi-agent systems with different $R_{rel}$ value

Following the study in Section 4.1, the two DDPG agents in the hand-shaking learning system share the same network parameters. To determine the unified setting for the relevance ratio in the multi-agent system, this paper investigated the performances of a PHEV controlled by the MADRL systems with the $R_{rel}$ value changing from 0 to 0.8 with 0.2 intervals. The simulation experiments were conducted under the learning cycles to investigate how the relevance ratio affects the multi-agent collaboration and the contributions of the learning agents to the global goal. And the learning curves of the two DDPG agents illustrating the evolution of agent rewards in 80 episodes are compared in Fig. 10.

The two agents are growing in opposite directions in the system with a $R_{rel}$ value of 0.4. In the system with a $R_{rel}$ value of 0.8, although both agents have the same tendency, the reward values of the multi-agent system decrease over time. Most of the studied cases ($R_{rel}$ = 0, 0.2, and 0.6) demonstrate the way of handshaking in the multi-agent system, in which both agents have the same tendency to grow with their reward values increase over time. Then, when $R_{rel}$ value of 0, 0.2, and 0.6, two agents perform well and achieve fast convergence at about 40 episodes and have the same tendency to reach the common goal. Therefore, the learning performance with more training time has been investigated in this paper.

The learning performance of the two agents with relevance ratio of 0, 0.2, and 0.6 are compared in Fig. 11. The result suggested that when the two agents have a relevant high level of relevance, i.e., $R_{rel}$ =0.6, although they follow the same trend during the training, both tend to be unstable and thus cannot converge at the steady point. By comparing the average value and standard deviation of the rewards obtained during the training with $R_{rel}$ values of 0 and 0.2 in Table 7, the study suggested that the multi-agent system with a $R_{rel}$ value of 0.2 is the best since it can achieve the highest average reward with less level of variation.

*4.2. Performance comparison with different methods*

To demonstrate the advantage of the proposed multi-agent system, a comparison study was conducted by using the rule-based method, the equivalent consumption minimization strategy (ECMS), and a single-agent system as baselines. By testing the PHEV under three driving cycles, the vehicle performances including the battery SoC at the end of the driving cycle, SoC sustaining error, fuel economy, and fuel saving rate over the rule-based method were obtained and compared in Table 8. The configuration of the single-agent system is summarized in Section 3, and the DDPG algorithm with the same setting as the multi-agent system is used for the online optimization of the single-agent system. By defining the battery SoC sustaining error, $SOC_{error}$, and fuel saving rate, $Saving$, as:

$$SOC_{error} = \frac{|SOC_{End} - SOC_{Initial}|}{SOC_{Initial}} * 100\% \qquad (28)$$

$$Saving = \frac{|Fuel_{Multi-agent} - Fuel_{Single-agent}|}{Fuel_{Single-agent}} * 100\%$$

where $SOC_{initial}$ and $SOC_{End}$ are the battery SoC value at the beginning and the end of a driving cycle, respectively; $Fuel_{single-agent}$ and $Fuel_{Multi-agent}$ are the fuel consumption in L/100 km for the single-agent system and the multi-agent system, respectively. Vehicle performances including fuel economy and battery SoC sustaining error are monitored under the learning cycle and two worldwide driving cycles, including UDDS and WLTC in Table 8 (a), (b), and (c), respectively.

Among the two traditional methods, the rule-based approach exhibits poorer performance in terms of both SoC sustaining and fuel economy. On the other hand, the ECMS method lacks stability for the testing environments with different initial SoC settings, limiting its generalization when compared to the single-agent system and the proposed multi-agent system. Notably, the ECMS methods demonstrated inconsistent performance in fuel economy and SoC sustaining error. For instance, when the initial SoC is set at 0.28 for the learning driving cycle, ECMS achieved very good fuel economy, however, the SoC sustaining error exceeded 5%, which is a performance threshold defined in legislation.

The results indicate that the multi-agent system can help maintain the battery SoC error within 5% when the initial SoC is relatively low (below 0.30). This means the PHEV controlled by the multi-agent system aligns with the requirement of most vehicle testing regulations around the world. The multi-agent system outperforms the single-agent system in all the study cases in terms of using less fuel and battery energy. This is because the multi-agent system generates two control variables to allow MG1 and MG2 to be controlled independently for their best performance. The PHEV controlled by the proposed multi-agent system achieved the best fuel economy in the learning driving cycle with all preset initial battery SoC values. Notably, significant fuel savings of up to 23.54% were achieved when the initial battery SoC is set at 0.25. Comparatively, in all driving cycles, the fuel economy performance is the most notable when the initial SoC is set at 0.25. Specifically, when the vehicle is driven under the WLTC cycle with an initial SoC of 0.25, the multi-agent system achieves fuel savings of 16.73% with a battery SoC sustaining error of only 1.60%, which far exceeded the industry standard (5%).

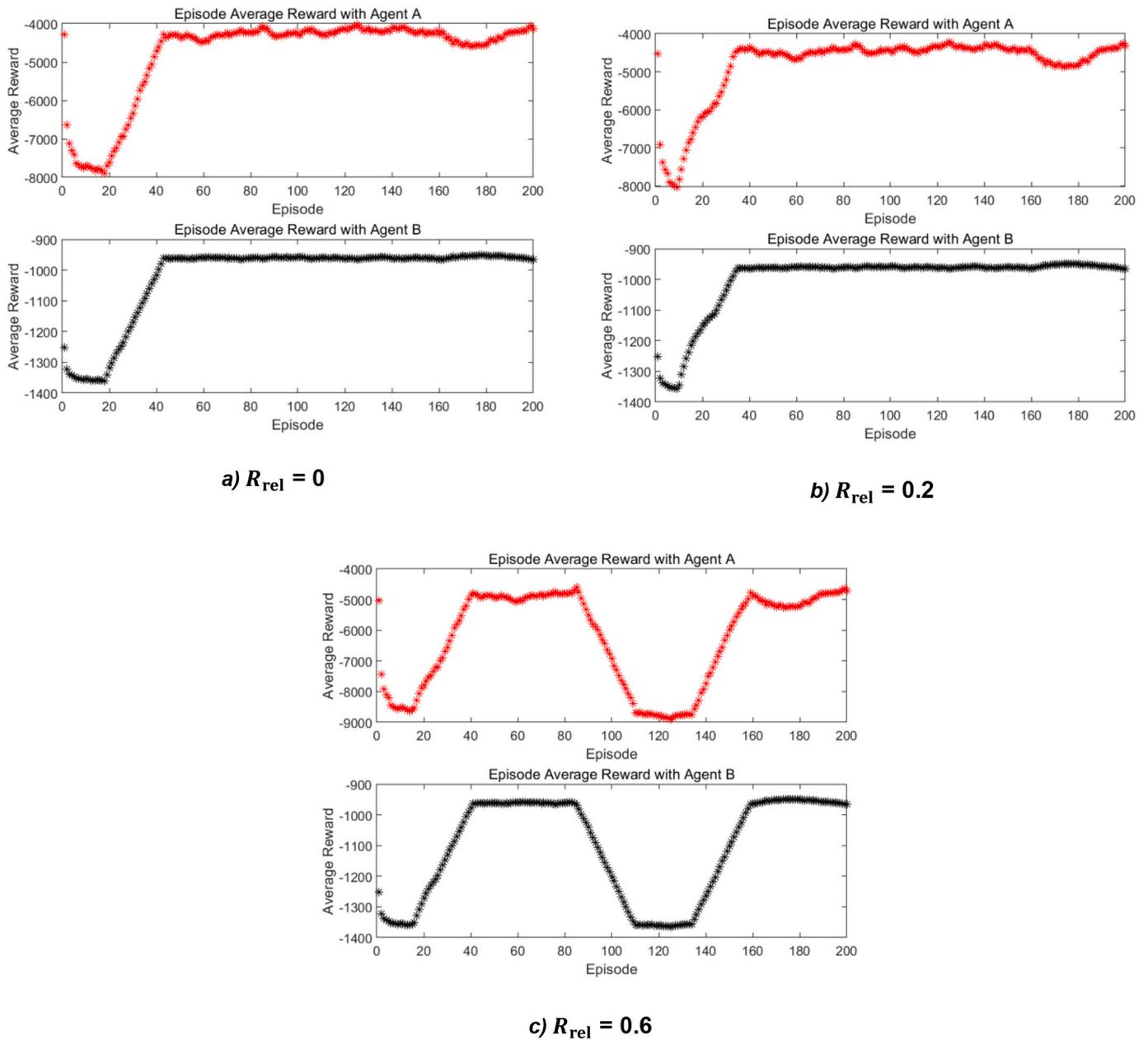

**Fig. 11.** Learning performance of a) $R_{\text{rel}} = 0$; b) $R_{\text{rel}} = 0.2$; c) $R_{\text{rel}} = 0.6$.

**Table 7**
Average values and standard deviation.

| $R_{rel}$ | Agent | Average reward value (All 200 episodes) | Standard deviation (40–200 episodes) |
|---|---|---|---|
| 0 | Agent A | −4782.1 | 48.064 |
|   | Agent B | −1018.8 | 48.064 |
| 0.2 | Agent A | −4803.3 | 46.621 |
|   | Agent B | −999.1 | 46.621 |

## 5. Conclusions

This paper studied a new MIMO control method for the multi-mode PHEV based on MADRL. By introducing a relevance ratio, a handshaking strategy has been proposed to enable two learning agents to work collaboratively under the MADRL framework using the DDPG algorithm. Through sensitivity analysis, parametric study, and software-in-the-loop testing, the conclusions drawn from the investigation are as follows:

**Table 8**
Learning performance of different methods.

(a) Learning driving cycle

| Initial SoC | Method | End SoC | SOC Error | Fuel (L/100 km) | Saving |
|---|---|---|---|---|---|
| 0.25 | Rule-based | 0.292 | 16.72% | 5.779 | – |
|  | ECMS | 0.280 | 12.04% | 5.625 | 2.67% |
|  | Single-agent | 0.272 | 8.80% | 4.534 | 21.55% |
|  | **Multi-agent** | **0.241** | **3.60%** | **4.419** | **23.54%** |
| 0.28 | Rule-based | 0.306 | 9.11% | 5.357 | – |
|  | ECMS | 0.280 | 0.04% | 4.598 | 14.17% |
|  | Single-agent | 0.305 | 8.93% | 4.547 | 15.12% |
|  | **Multi-agent** | **0.271** | **3.21%** | **4.450** | **16.93%** |
| 0.3 | Rule-based | 0.312 | 4.10% | 5.027 | – |
|  | ECMS | 0.280 | 6.63% | 3.926 | 21.89% |
|  | Single-agent | 0.328 | 9.33% | 4.534 | 9.81% |
|  | **Multi-agent** | **0.286** | **4.67%** | **4.418** | **12.11%** |

(b) UDDS driving cycle

| Initial SoC | Method | End SoC | SOC Error | Fuel (L/100 km) | Saving |
|---|---|---|---|---|---|
| 0.25 | Rule-based | 0.307 | 22.88% | 5.368 | – |
|  | ECMS | 0.298 | 19.36% | 5.077 | 5.42% |
|  | Single-agent | 0.262 | 4.80% | 4.43 | 17.47% |
|  | **Multi-agent** | **0.261** | **4.40%** | **4.351** | **18.95%** |
| 0.28 | Rule-based | 0.322 | 14.93% | 4.916 | – |
|  | ECMS | 0.298 | 6.57% | 3.899 | 20.69% |
|  | Single-agent | 0.292 | 4.29% | 4.43 | 9.89% |
|  | **Multi-agent** | **0.29** | **3.57%** | **4.306** | **12.41%** |
| 0.3 | Rule-based | 0.330 | 10.00% | 4.573 | – |
|  | ECMS | 0.298 | 0.53% | 4.423 | 3.27% |
|  | Single-agent | 0.312 | 4.00% | 4.431 | 3.10% |
|  | **Multi-agent** | **0.310** | **3.33%** | **4.327** | **5.37%** |

(c) WLTP driving cycle

| Initial SoC | Method | End SoC | SOC Error | Fuel (L/100 km) | Saving |
|---|---|---|---|---|---|
| 0.25 | Rule-based | 0.315 | 26.08% | 4.990 | – |
|  | ECMS | 0.281 | 12.44% | 4.518 | 9.45% |
|  | Single-agent | 0.261 | 4.40% | 4.324 | 13.34% |
|  | **Multi-agent** | **0.246** | **1.60%** | **4.155** | **16.73%** |
| 0.28 | Rule-based | 0.321 | 14.50% | 4.640 | – |
|  | ECMS | 0.281 | 0.40% | 4.205 | 9.38% |
|  | Single-agent | 0.293 | 4.64% | 4.324 | 6.82% |
|  | **Multi-agent** | **0.292** | **4.29%** | **4.148** | **10.61%** |
| 0.3 | Rule-based | 0.324 | 7.90% | 4.413 | – |
|  | ECMS | 0.281 | 6.30% | 3.501 | 20.67% |
|  | Single-agent | 0.314 | 4.67% | 4.325 | 1.99% |
|  | **Multi-agent** | **0.304** | **1.33%** | **4.162** | **5.68%** |

- represents the baseline performance.

1) The learning rate of the DDPG agents in the proposed MADRL-based EMS is the most significant influencing factor in determining the learning performance including computing time, the number of episodes for convergence, and fuel economy.
2) Hand-shaking among the DDPG agents is achievable by tuning the relevance ratio. The optimal setting for the relevance ratio is 0.2 for control of the studied multi-mode PHEV. A smaller relevance ratio would lead to a low learning speed while a higher value will make the system unstable.
3) The proposed MADRL method outperforms the baseline methods (single agent learning method, rule-based method, and ECMS) under all studied driving cycles in terms of mitigating energy consumption and battery SoC sustaining error. Up to 23.54% of fuel can be saved with battery SoC error well-controlled within 5% compared to the baseline methods.

Low-latency decision-making algorithms and real-time coordination between multiple agents will be studied and validated in the planned future work with real-time hardware-in-the-loop facilities. This aims to make contributions to bridging the gap between simulation-based results and real-world deployment, enabling its potential application in commercial PHEVs.

**CRediT authorship contribution statement**

**Min Hua:** Conceptualization, Formal analysis, Methodology, Project administration, Software, Writing – original draft. **Cetengfei Zhang:** Formal analysis, Validation. **Fanggang Zhang:** Methodology, Software, Validation. **Zhi Li:** Conceptualization, Methodology, Project administration, Writing – review & editing. **Xiaoli Yu:** Funding acquisition, Project administration, Supervision. **Hongming Xu:** Supervision, Writing – review & editing. **Quan Zhou:** Conceptualization, Funding acquisition, Methodology, Supervision, Writing – review & editing.

**Declaration of Competing Interest**

The authors declare the following financial interests/personal relationships which may be considered as potential competing interests: Prof Hongming Xu is an associate editor of Applied Energy.

**Data availability**

Data will be made available on request.

**Acknowledgment**

This work was supported by the State Key Laboratory of Clean Energy Utilization (Open Fund Project, No. ZJUCEU2021019).